\documentclass[11pt,a4paper]{article}
\usepackage[hyperref]{emnlp2020}
\usepackage{times}
\usepackage{latexsym}

\usepackage{fancyhdr,graphicx}
\usepackage{multirow}
\usepackage{subcaption}

\usepackage[utf8x]{inputenc}
\usepackage[thaicjk,english]{babel}
\addto\extrasthaicjk{\fontencoding{C90}\selectfont}

\makeatletter
\@namedef{opt@inputenc.sty}{utf8}
\makeatother
\usepackage{CJKutf8}

\usepackage{microtype}

\aclfinalcopy 

\title{Cross-lingual Spoken Language Understanding with \\ Regularized Representation Alignment}

\author{Zihan Liu, Genta Indra Winata, Peng Xu, Zhaojiang Lin, Pascale Fung \\
Center for Artificial Intelligence Research (CAiRE)\\
Department of Electronic and Computer Engineering\\
The Hong Kong University of Science and Technology, Clear Water Bay, Hong Kong\\
\texttt{zihan.liu@connect.ust.hk}, \texttt{pascale@ece.ust.hk}}

\date{}

\begin{document}
\maketitle
\begin{abstract}

Despite the promising results of current cross-lingual models for spoken language understanding systems, they still suffer from imperfect cross-lingual representation alignments between the source and target languages, which makes the performance sub-optimal.
To cope with this issue, we propose a regularization approach to further align word-level and sentence-level representations across languages without any external resource.
First, we regularize the representation of user utterances based on their corresponding labels.
Second, we regularize the latent variable model~\cite{liu2019zero} by leveraging adversarial training to disentangle the latent variables.
Experiments on the cross-lingual spoken language understanding task show that our model outperforms current state-of-the-art methods in both few-shot and zero-shot scenarios, and our model, trained on a few-shot setting with only 3\% of the target language training data, achieves comparable performance to the supervised training with all the training data.\footnote{The code is available in \url{https://github.com/zliucr/crosslingual-slu}.}

\end{abstract}

\section{Introduction}
Data-driven neural-based supervised training approaches have shown effectiveness in spoken language understanding (SLU) systems~\cite{goo2018slot,chen2019bert,haihong2019novel}. However, collecting large amounts of high-quality training data is not only expensive but also time-consuming, which makes these approaches not scalable to low-resource languages due to the scarcity of training data. Cross-lingual adaptation has naturally arisen to cope with this issue, which leverages the training data in rich-resource source languages and minimizes the requirement of training data in low-resource target languages.

\begin{figure}[!ht]
    \centering
    \includegraphics[scale=0.86]{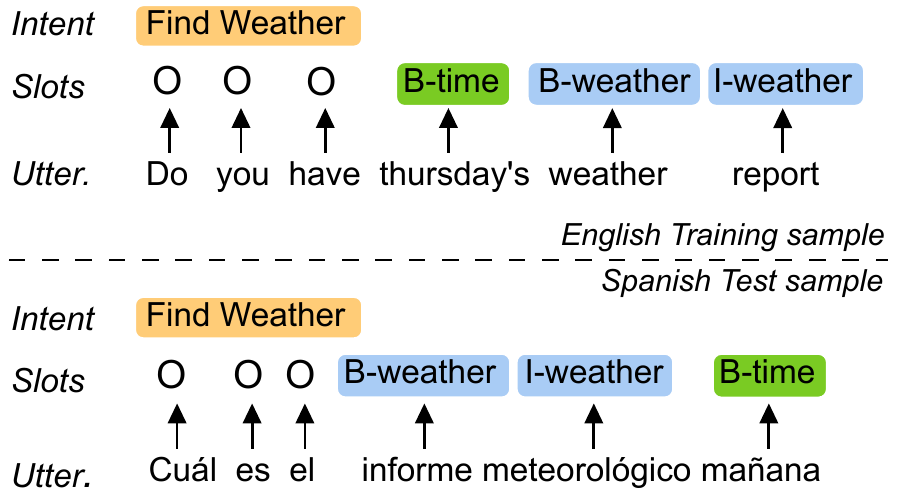}
    \caption{Illustration of cross-lingual spoken language understanding systems, where English is the source language and Spanish is the target language.}
    \label{fig:example}
\end{figure}



In general, there are two challenges in cross-lingual adaptation. First, the imperfect alignment of word-level representations between the source and target language limits the adaptation performance. Second, even though we assume that the word-level alignment is perfect, the sentence-level alignment is still imperfect owing to grammatical and syntactical variances across languages. 
Therefore, we emphasize that cross-lingual methods should focus on the alignments of word-level and sentence-level representations, and increase the robustness for inherent imperfect alignments.

In this paper, we concentrate on the cross-lingual SLU task (as illustrated in Figure~\ref{fig:example}),
and we consider both few-shot and zero-shot scenarios.
To improve the quality of cross-lingual alignment, we first propose a \textbf{L}abel \textbf{R}egularization (\textbf{LR}) method, which utilizes the slot label sequences to regularize the utterance representations. 
We hypothesize that if the slot label sequences of user utterances are close to each other, these user utterances should have similar meanings. Hence, we regularize the distance of utterance representations based on the corresponding representations of label sequences to further improve the cross-lingual alignments.


Then, we extend the latent variable model (LVM) proposed by~\citet{liu2019zero}. The LVM generates a Gaussian distribution instead of a feature vector for each token, which improves the adaptation robustness. However, there are no additional constraints on generating distributions, making the latent variables easily entangled for different slot labels. To handle this issue, we leverage \textbf{A}dversarial training to regularize the \textbf{LVM} (\textbf{ALVM}). 
We train a linear layer to fit latent variables to a uniform distribution over slot types. Then, we optimize the latent variables to fool the trained linear layer to output the correct slot type (one hot vector). In this way, latent variables of different slot types are encouraged to disentangle from each other, leading to a better alignment of cross-lingual representations. 

The contributions of our work are summarized as follows:
\begin{itemize}
    \item We propose LR and ALVM to further improve the alignment of cross-lingual representations, which do not require any external resources.
    \item Our model outperforms the previous state-of-the-art model in both zero-shot and few-shot scenarios on the cross-lingual SLU task.
    \item Extensive analysis and visualizations are made to illustrate the effectiveness of our approaches.
\end{itemize}



\begin{figure*}[!ht]
    \centering
    \resizebox{1.0\textwidth}{!}{
    \includegraphics{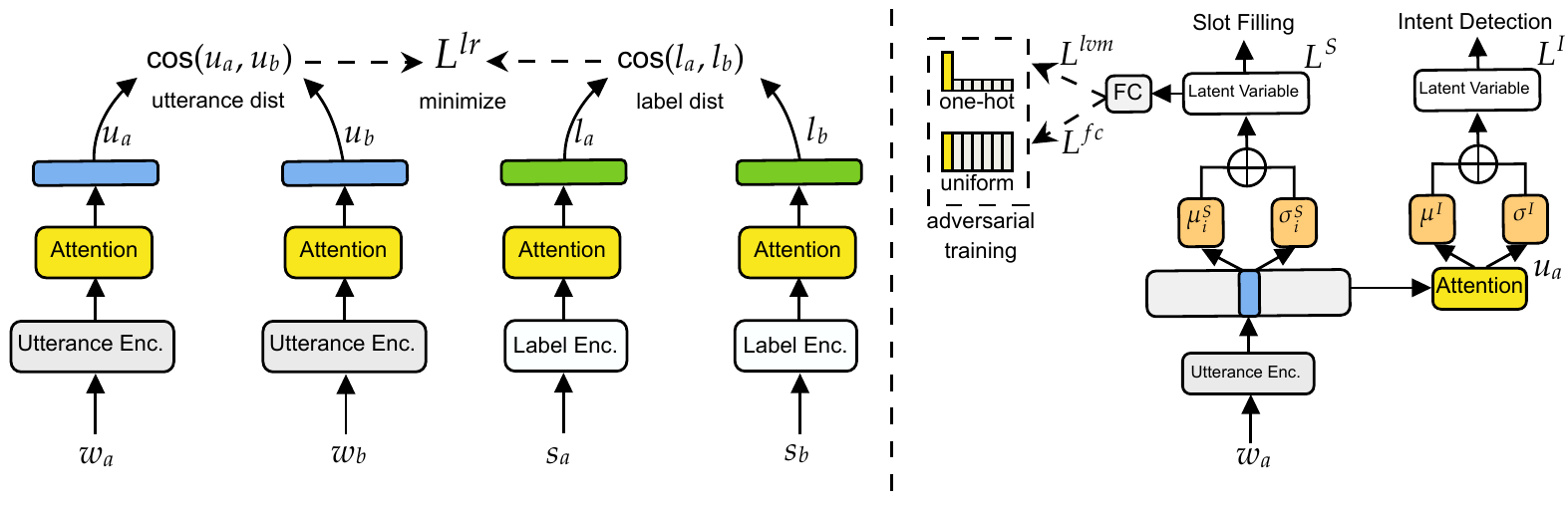}
    }
    \caption{Left: Illustration of label regularization (LR). 
    Right: The model architecture with adversarial latent variable model (ALVM), where $FC$ consists of a linear layer and a softmax function.}
    \label{fig:model}
\end{figure*}

\section{Related Work}
\paragraph{Cross-lingual Transfer Learning}
Cross-lingual transfer learning is able to circumvent the requirement of enormous training data by leveraging the learned knowledge in the source language and learning inter-connections between the source and the target language. \citet{artetxe2017learning} and \citet{conneau2017word} conducted cross-lingual word embedding mapping with zero or very few supervision signals. Recently, pre-training cross-lingual language models on large amounts of monolingual or bilingual resources have been proved to be effective for the downstream tasks (e.g., natural language inference)~\cite{lample2019cross, devlin2019bert,pires-etal-2019-multilingual,huang2019unicoder}. Additionally, many cross-lingual transfer algorithms have been proposed to solve specific cross-lingual tasks, for example, \textit{named entity recognition}~\cite{xie2018neural, mayhew2017cheap, liu2020exploring}, \textit{part of speech tagging}~\cite{kim2017cross, zhang2016ten}, \textit{entity linking}~\cite{zhang2013cross,sil2018neural,upadhyay2018joint}, personalized conversations~\cite{lin2020xpersona},
and dialog systems~\cite{upadhyay2018almost,chen2018xl}.

\paragraph{Cross-lingual Task-oriented Dialog Systems}
Deploying task-oriented dialogue systems in low-resource domains~\cite{bapna2017towards,wu2019transferable,liu-etal-2020-coach} or languages~\cite{chen2018xl,liu2019zero,liu2019attention}, where the number of training of samples is limited, is a challenging task. \citet{mrkvsic2017semantic} expanded Wizard of Oz (WOZ) into multilingual WOZ by annotating two additional languages.
\citet{schuster2019cross} introduced a multilingual SLU dataset and proposed to leverage bilingual corpus and multilingual CoVe~\cite{yu-etal-2018-multilingual} to align the representations across languages. \citet{chen2018xl} proposed a teacher-student framework based on a bilingual dictionary or bilingual corpus for building cross-lingual dialog state tracking. Instead of highly relying on extensive bilingual resources, \citet{qin2020cosda} introduced a data augmentation framework to generate multilingual code-switching data for cross-lingual tasks including the SLU task. \citet{liu2019attention} leveraged a mixed language training framework for cross-lingual task-oriented dialogue systems. And \citet{liu2019zero} proposed to refine the cross-lingual word embeddings by using very few word pairs, and introduced a latent variable model to improve the robustness of zero-shot cross-lingual SLU. Nevertheless, there still exists improvement space for the cross-lingual alignment. In this paper, we propose to further align the cross-lingual representations so as to boost the performance of cross-lingual SLU systems.

\section{Methodology}
Our model architecture and proposed methods are depicted in Figure~\ref{fig:model}, and combine label regularization (LR) and the adversarial latent variable model (ALVM) to conduct the intent detection and slot filling.
In the few-shot setting, the input user utterances are in both the source and target languages, while in the zero-shot setting, the user utterances are only in the source language. Note that both the source and target languages contain only one language.

\subsection{Label Regularization}

\subsubsection{Motivation}
Intuitively, when the slot label sequences are similar, we expect the corresponding representations of user utterances across languages to be similar. For example, when the slot label sequences contain the weather slot and the location slot, the user utterances should be asking for the weather forecast somewhere. 
However, the representations of utterances across languages can not always meet these requirements because of the inherent imperfect alignments in word-level and sentence-level representations.
Therefore, we propose to leverage existing slot label sequences in the training data to regularize the distance of utterance representations. 

When a few training samples are available in the target language (i.e., few-shot setting), we regularize the distance of utterance representations between the source and target languages based on their slot labels. Given this regularization, the model explicitly learns to further align the sentence-level utterance representations across languages so as to satisfy the constraints. Additionally, it can also implicitly align the word-level BiLSTM hidden states across languages because sentence-level representations are produced based on them.

When zero training samples are available in the target language (i.e., zero-shot setting), we regularize the utterance representations in the source language. It can help better distinguish the utterance representations and cluster similar utterance representations based on the slot labels, which increases the generalization ability in the target language.

\subsubsection{Implementation Details}
Figure~\ref{fig:model} (Left) illustrates an utterance encoder and a label encoder that generate the representations for utterances and labels, respectively. 

We denote the user utterance as $ \textbf{w} = [w_1, w_2, ..., w_{n}] $, where $n$ is the length of the utterance.
Similarly, we represent the slot label sequences as $ \textbf{s} = [s_1, s_2, ..., s_{n}] $.
We combine a bidirectional LSTM (BiLSTM)~\cite{hochreiter1997long} and an attention layer~\cite{felbo2017using} to encode and produce the representations for user utterances and slot label sequences. The representation generation process is defined as follows:
\begin{equation}
    [h_1^{w}, h_2^{w}, ..., h_{w}^{w}] = \textnormal{BiLSTM}_{\textnormal{utter}} (\textbf{E}(\textbf{w})),
\end{equation}
\begin{equation}
    [h_1^{s}, h_2^{s}, ..., h_{n}^{s}] = \textnormal{BiLSTM}_{\textnormal{label}} (\textbf{E}(\textbf{s})),
\end{equation}
\begin{equation}
    m_i^{w} = h_i^{w} v^{w},~~ \alpha_i^{w} = \frac{exp(m_i^{w})}{\sum_{t=1}^{n}exp(m_t^{w})},
\end{equation}
\begin{equation}
    m_i^{s} = h_i^{s} v^{s},~~ \alpha_i^{s} = \frac{exp(m_i^{s})}{\sum_{t=1}^{n}exp(m_t^{s})},
\end{equation}
\begin{equation}
    u = \sum_{i=1}^{n} \alpha_i^{w} h_i^{w},~~ l = \sum_{i=1}^{n} \alpha_i^{s} h_i^{s},
\end{equation}
where the superscript $w$ and $s$ represents utterance and label, respectively, $v$ is a trainable weight vector in the attention layer, $\alpha_i$ is the attention score for each token i, \textbf{E} denotes the embedding layers for utterances and label sequences, and $u$ and $l$ denotes the representation of utterance $\textbf{w}$ and slot label $\textbf{s}$, respectively.

In each iteration of the training phase, we randomly select two samples for the label regularization. As illustrated in Figure~\ref{fig:model} (Left),
we first calculate the cosine similarity of two utterance representations $u_a$ and $u_b$, and the cosine similarity of two label representations $l_a$ and $l_b$. Then, we minimize the distance of these two cosine similarities. The objective functions 
can be described as follows:
\begin{equation}
    \textnormal{cos}(u_a,u_b) = \frac{u_{a} \cdot u_{b}}{||u_{a}||~||u_{b}||},
\end{equation}
\begin{equation}
    \textnormal{cos}(l_a,l_b) = \frac{l_{a} \cdot l_{b}}{||l_{a}||~||l_{b}||}, 
\end{equation}
\begin{equation}
    L^{lr} = \sum_{a,b} \textnormal{MSE}(\textnormal{cos}(u_a,u_b), \textnormal{cos}(l_a,l_b)), \label{eq8}
\end{equation}
where the superscript $lr$ denotes label regularization, and \textnormal{MSE} represents mean square error.
In the zero-shot setting, both samples $u_a$ and $u_b$ come from the source language. While in the few-shot setting, one sample comes from the source language and the other one comes from the target language.

Since the features of labels and utterances are in different vector spaces, we choose not to share the parameters of their encoders. During training, it is easy to produce expressive representations for user utterances due to the large training samples, but it is difficult for label sequences since the objective function $L^{lr}$ is the only supervision. This supervision is weak at the beginning of the training since utterance representations are not sufficiently expressive, which leads to the label regularization approach not being stable and effective.
To ensure the representations for slot label sequences are meaningful, we conduct pre-training for the label sequence encoder.

\subsubsection{Label Sequence Encoder Pre-training}
We leverage the large amount of source language training data to pre-train the label sequence encoder. Concretely, we use the model architecture illustrated in Figure~\ref{fig:model} to train the SLU system in the source language, and at the same time, we optimize the label sequence encoder based on the objective function $L^{lr}$ in Eq~(\ref{eq8}). The label sequence encoder learns to generate meaningful label sequence representations that differ based on their similarities since the extensive source language training samples ensure the high quality of the utterance encoder. 



\subsection{Adversarial Latent Variable Model}
In this section, we first give an introduction to the latent variable model (LVM)~\cite{liu2019zero}, and then we describe how we incorporate the adversarial training into the LVM.

\subsubsection{Latent Variable Model}
Point estimation in the cross-lingual adaptation is vulnerable due to the imperfect alignments across languages. Hence, as illustrated in Figure~\ref{fig:model} (Right), the LVM generates a Gaussian distribution with mean $\mu$ and variance $\sigma$ for both word-level and sentence-level representations instead of a feature vector, which eventually improves the robustness of the model's cross-lingual adaptation ability. The LVM can be formulated as
\begin{equation}
    \left[ \begin{array}{c} { \mu^S_i } \\ { \log (\sigma^S_i)^{2} } \end{array} \right] = \mathbf{W}^S_l h_i, \left[ \begin{array} { c } { \mu^I } \\  { \log (\sigma^I)^{2} } \end{array} \right] = \mathbf{W}^I_l u,
\end{equation}
\begin{equation}
    z^S_i \sim q^S_i (z | h_i), ~~ z^I \sim q^I (z | u),
\end{equation}
\begin{equation}
    p^S_i (s_i | z^S_i) = \textnormal{Softmax}(\mathbf{W}^S_p z^S_i),
\end{equation}
\begin{equation}
    p^I (I | z^I) = \textnormal{Softmax}(\mathbf{W}^I_p z^I),
\end{equation}
where $W_{l}^S$ and $W_{l}^I$ are trainable parameters to generate the mean and variance for word-level hidden states $h_i$ and sentence-level representations $r$, respectively, from user utterances. $q^S_i \sim \mathcal{N}(\mu^S_i, (\sigma^S_i)^2\mathbf{I})$ and $q^I \sim \mathcal{N}(\mu^I, (\sigma^I)^2\mathbf{I})$ are the generated Gaussian distributions, which latent variables $z^S_t$ and $z^I$ are sampled from, and $p^S_i$ and $p^I$ is the predictions for the slot of the $i^{th}$ token and the intent of the utterance, respectively. 

During training, all the sampled points from the same generated distribution will be trained to predict the same slot label, which makes the adaptation more robust. In the inference time, the true mean $\mu^S_i$ and $\mu^I$ is used to replace $z^S_i$ and $z^I$, respectively, to make the prediction deterministic.

\subsubsection{Adversarial Training}

Since there are no constraints enforced on the latent Gaussian distribution during training, the latent distributions of different slot types are likely to be close to each other.
Hence, the distributions for the same slot type in different user utterances or languages might not be clustered well,
which could hurt the cross-lingual alignment and prevent the model from distinguishing slot types when adapting to the target language.

To improve the cross-lingual alignment of latent variables, we propose to make the latent variables of different slot types more distinguishable by adding adversarial training to the LVM. As illustrated in Figure~\ref{fig:model} (Right), we train a fully connected layer to fit latent variables into a uniform distribution over slot types. At the same time, the latent variables are regularized to fool the trained fully connected layer by predicting the correct slot type. 
In this way, the latent variables are trained to be more recognizable. In other words, the generated distributions for different slot types are more likely to repel each other, and for the same slot type are more likely to be close to each other, which leads to a more robust cross-lingual adaptation.
We denote the size of the whole training data as $J$ and the length for data sample $j$ as $|Y_j|$. Note that in the few-shot setting, $J$ includes the number of data samples in the target language.
The process of adversarial training can be described as follows:
\begin{equation}
    p_{jk} = \mathcal{FC}(z_{jk}^S),
\end{equation}
\begin{equation}
    L^{fc} = \sum_{j=1}^J \sum_{k=1}^{|Y_j|} \textnormal{MSE}(p_{jk}, \mathcal{U}),
\end{equation}
\begin{equation}
    L^{lvm} = \sum_{j=1}^J \sum_{k=1}^{|Y_j|} \textnormal{MSE}(p_{jk}, y_{jk}^{S}),
\end{equation}
where $\mathcal{FC}$ consists of a linear layer and a \textnormal{Softmax} function, and $z_{jk}^{S}$ and $p_{jk}$ is the latent variable and generated distribution, respectively, for the $k^{th}$ token in the $j^{th}$ utterance, 
$\textnormal{MSE}$ represents the mean square error,  $\mathcal{U}$ represents the uniform distribution, and $y_{jk}^{S}$ represents the slot label. The slot label is a one-hot vector where the value for the correct slot type is one and zero otherwise.
We optimize $L^{fc}$ to train only $\mathcal{FC}$ to fit a uniform distribution, and $L^{lvm}$ is optimized to constrain the LVM to generate more distinguishable distributions for slot predictions.
Different from the well-known adversarial training~\cite{goodfellow2014generative} where the discriminator is to distinguish the classes, and the generator is to make the features not distinguishable, in our approach, the $\mathcal{FC}$ layer, acting as the discriminator, is trained to generate uniform distribution, and the generator is regularized to make latent variables distinguishable by slot types.

\subsection{Optimization}
The objective functions for the slot filling and intent detection tasks are illustrated as follows:
\begin{equation}
    L^S = \sum_{j=1}^{J} \sum_{k=1}^{|Y_j|} -log(p_{jk}^{S} \cdot (y_{jk}^{S})^\top),
\end{equation}
\begin{equation}
    L^I = \sum_{j=1}^{J} -log(p_{j}^{I} \cdot (y_{j}^I)^\top),
\end{equation}
where $p_{jk}^S$ and $y_{jk}^S$ is the prediction and label, respectively, for the slot of the $k^{th}$ token in the $j^{th}$ utterance, and $p_{j}^{I}$ and $y_{j}^{I}$ is the intent prediction and label, respectively, for the $j^{th}$ utterance.

The optimization for our model is to minimize the following objective function:
\begin{equation}
    L = L^S + L^I + L^{lr} + \alpha L^{fc} + \beta L^{lvm}, \label{eq19}
\end{equation}
where $\alpha$ and $\beta$ are hyper-parameters,
$L^{fc}$ only optimizes the parameters in $\mathcal{FC}$, and $L^{lvm}$ optimizes all the model parameters excluding $\mathcal{FC}$.

\section{Experiments}
\subsection{Dataset}
We conduct our experiments on the multilingual spoken language understanding (SLU) dataset proposed by~\citet{schuster2019cross}, which contains English, Spanish, and Thai across the weather, reminder, and alarm domains. The corpus includes 12 intent types and 11 slot types, and the data statistics are shown in Table~\ref{table:data_statistics}.

\begin{table}[]
\centering
\resizebox{0.41\textwidth}{!}{
\begin{tabular}{l|c|c|c}
\hline
\textbf{\# Utterance} & \textbf{English}  & \textbf{Spanish}  & \textbf{Thai}  \\ \hline
Train & 30,521 & 3,617 & 2,156 \\
Validation & 4,181 & 1,983 & 1,235 \\
Test & 8,621 & 3,043 & 1,692 \\ \hline
\end{tabular}
}
\caption{Number of utterances for the multilingual SLU dataset. English is the source language, and Spanish and Thai are the target languages.}
\label{table:data_statistics}
\end{table}

\begin{table*}[!t]
\centering
\resizebox{0.99\textwidth}{!}{
\begin{tabular}{lcccccccc}
\hline
\multicolumn{1}{l|}{\multirow{2}{*}{\textbf{Model}}} & \multicolumn{4}{c|}{\textbf{Spanish}} & \multicolumn{4}{c}{\textbf{Thai}} \\ \cline{2-9} 
\multicolumn{1}{l|}{} & \multicolumn{2}{c}{\textbf{Intent Acc.}} & \multicolumn{2}{|c|}{\textbf{Slot F1}} & \multicolumn{2}{c|}{\textbf{Intent Acc.}} & \multicolumn{2}{c}{\textbf{Slot F1}} \\ \hline \hline
\multicolumn{9}{l}{\textbf{Few-shot settings}} \\ \hline
\multicolumn{1}{l|}{} & $\tt{1\%}$-$\tt{shot}$ & \multicolumn{1}{c|}{$\tt{3\%}$-$\tt{shot}$} & $\tt{1\%}$-$\tt{shot}$ & \multicolumn{1}{c|}{$\tt{3\%}$-$\tt{shot}$} & $\tt{1\%}$-$\tt{shot}$ & \multicolumn{1}{c|}{$\tt{3\%}$-$\tt{shot}$} & $\tt{1\%}$-$\tt{shot}$ & $\tt{3\%}$-$\tt{shot}$ \\ \hline
\multicolumn{1}{l|}{BiLSTM-CRF} & 93.03 & \multicolumn{1}{c|}{93.63} & 75.70 & \multicolumn{1}{c|}{82.60} & 81.30 & \multicolumn{1}{c|}{87.23} & 52.57 & 66.04 \\
\multicolumn{1}{l|}{\hspace{1mm}+ LR} & 93.08 & \multicolumn{1}{c|}{95.04} & 77.04 & \multicolumn{1}{c|}{84.09} & 84.04 & \multicolumn{1}{c|}{89.20} & 57.40 & 67.45 \\  \hline
\multicolumn{1}{l|}{BiLSTM-LVM} & 92.86 & \multicolumn{1}{c|}{94.46} & 75.19 & \multicolumn{1}{c|}{82.64} & 83.51 & \multicolumn{1}{c|}{89.08} & 55.08 & 67.26 \\
\multicolumn{1}{l|}{\hspace{1mm}+ LR} & 93.79 & \multicolumn{1}{c|}{95.16} & 76.96 & \multicolumn{1}{c|}{83.54} & 86.33 & \multicolumn{1}{c|}{90.80} & 59.02 & 70.26 \\
\multicolumn{1}{l|}{\hspace{1mm}+ ALVM} & 93.78 & \multicolumn{1}{c|}{95.27} & 78.35 & \multicolumn{1}{c|}{83.69} & 85.40 & \multicolumn{1}{c|}{90.70} & 59.75 & 69.38 \\
\multicolumn{1}{l|}{\hspace{1mm}+ LR \& ALVM} & 93.82 & \multicolumn{1}{c|}{95.20} & 78.46 & \multicolumn{1}{c|}{84.19} & 87.43 & \multicolumn{1}{c|}{90.96} & 61.44 & 70.88 \\
\multicolumn{1}{l|}{\hspace{1mm}+ LR \& ALVM \& delex.} & \textbf{94.71} & \multicolumn{1}{c|}{\textbf{95.62}} & \textbf{80.82} & \multicolumn{1}{c|}{\textbf{85.18}} & \textbf{87.67} & \multicolumn{1}{c|}{\textbf{91.61}} & \textbf{62.01} & \textbf{72.39}  \\ \hline
\multicolumn{1}{l|}{XL-SLU} & 92.70 & \multicolumn{1}{c|}{94.96} & 77.67 & \multicolumn{1}{c|}{82.22} & 84.04 & \multicolumn{1}{c|}{89.59} & 55.57 & 67.56 \\
\multicolumn{1}{l|}{M-BERT} & 92.77 & \multicolumn{1}{c|}{95.56} & 80.15 & \multicolumn{1}{c|}{84.50} & 83.87 & \multicolumn{1}{c|}{89.19} & 58.18 & 67.88 \\  \hline \hline
\multicolumn{9}{l}{\textbf{Zero-shot settings}} \\ \hline
\multicolumn{1}{l|}{XL-SLU} & \multicolumn{2}{c|}{90.20} & \multicolumn{2}{c|}{65.79} & \multicolumn{2}{c|}{73.43} & \multicolumn{2}{c}{32.24} \\
\multicolumn{1}{l|}{\hspace{1mm}+ LR} & \multicolumn{2}{c|}{91.51} & \multicolumn{2}{c|}{71.55} & \multicolumn{2}{c|}{74.86} & \multicolumn{2}{c}{32.86} \\
\multicolumn{1}{l|}{\hspace{1mm}+ ALVM} & \multicolumn{2}{c|}{91.48} & \multicolumn{2}{c|}{71.21} & \multicolumn{2}{c|}{74.35} & \multicolumn{2}{c}{32.97} \\
\multicolumn{1}{l|}{\hspace{1mm}+ LR \& ALVM} & \multicolumn{2}{c|}{\textbf{92.31}} & \multicolumn{2}{c|}{\textbf{72.49}} & \multicolumn{2}{c|}{\textbf{75.77}} & \multicolumn{2}{c}{\textbf{33.28}} \\ \hline
\multicolumn{1}{l|}{MLT} & \multicolumn{2}{c|}{86.54} & \multicolumn{2}{c|}{74.43} & \multicolumn{2}{c|}{70.57} & \multicolumn{2}{c}{28.47} \\
\multicolumn{1}{l|}{CoSDA-ML} & \multicolumn{2}{c|}{\textbf{94.80}} & \multicolumn{2}{c|}{\textbf{80.40}} & \multicolumn{2}{c|}{\textbf{76.80}} & \multicolumn{2}{c}{\textbf{37.3}} \\
\multicolumn{1}{l|}{M-BERT} & \multicolumn{2}{c|}{74.91} & \multicolumn{2}{c|}{67.55} & \multicolumn{2}{c|}{42.97} & \multicolumn{2}{c}{10.68} \\
\multicolumn{1}{l|}{Multi. CoVe} & \multicolumn{2}{c|}{53.34} & \multicolumn{2}{c|}{22.50} & \multicolumn{2}{c|}{66.35} & \multicolumn{2}{c}{32.52} \\
\multicolumn{1}{l|}{\hspace{1mm}+ Auto-encoder} & \multicolumn{2}{c|}{53.89} & \multicolumn{2}{c|}{19.25} & \multicolumn{2}{c|}{70.70} & \multicolumn{2}{c}{35.62} \\ \hline 
\multicolumn{1}{l|}{Translate Train} & \multicolumn{2}{c|}{\textit{85.39}} & \multicolumn{2}{c|}{\textit{72.89}} & \multicolumn{2}{c|}{\textit{95.89}} & \multicolumn{2}{c}{\textit{55.43}} \\ \hline \hline
\multicolumn{9}{l}{\textbf{All-shot settings}} \\ \hline
\multicolumn{1}{l|}{Target$^\dagger$} & \multicolumn{2}{c|}{\textit{96.08}} & \multicolumn{2}{c|}{\textit{86.03}} & \multicolumn{2}{c|}{\textit{92.73}} & \multicolumn{2}{c}{\textit{85.52}} \\
\multicolumn{1}{l|}{Source \& Target$^\ddagger$} & \multicolumn{2}{c|}{\textit{98.06}} & \multicolumn{2}{c|}{\textit{87.65}} & \multicolumn{2}{c|}{\textit{95.58}} & \multicolumn{2}{c}{\textit{88.11}} \\ \hline
\end{tabular}
}
\caption{Cross-lingual SLU results (averaged over three runs). $^\dagger$denotes supervised training on all the target language training samples. $^\ddagger$denotes supervised training on both the source and target language datasets. The bold numbers denote the best results in the few-shot or zero-shot settings. The underlined numbers represent that the results are comparable (distances are within 1\%) to the all-shot experiment with all the target language training samples. The results of Multi. CoVe and Multi. CoVe + Auto-encoder are taken from~\citet{schuster2019cross}, and the results of XL-SLU in the zero-shot settings are taken from~\citet{liu2019zero}.}
\label{table:few-zero-shot}
\end{table*}

\subsection{Training Details}
The utterance encoder is a 2-layer BiLSTM with a hidden size of 250 and dropout rate of 0.1, and the size of the mean and variance in the latent variable model is 150. The label encoder is a 1-layer BiLSTM with a hidden size of 150, and 100-dimensional embeddings for label types. We use the Adam optimizer with a learning rate of 0.001. 
We use \textit{accuracy} to evaluate the performance of intent detection and BIO-based \textit{f1-score} to evaluate the performance of slot filling. 
For the adversarial training, we realize that the latent variable model is not able to make slot types recognizable if the $\mathcal{FC}$ is too strong.
Hence, we decide to first learn a good initialization for $\mathcal{FC}$ by setting both $\alpha$ and $\beta$ parameters in Eq~(\ref{eq19}) as 1 in the first two training epochs, and then we gradually decrease the value of $\alpha$. 
We use the refined cross-lingual word embeddings in~\citet{liu2019zero}~\footnote{Available at https://github.com/zliucr/Crosslingual-NLU} to initialize the cross-lingual word embeddings in our models and let them not be trainable.
We use the delexicalization (delex.) in~\citet{liu2019zero}, which replaces the tokens that represent numbers, time, and duration with special tokens. We use \textbf{36} training samples in Spanish and \textbf{21} training samples in Thai on the 1\% few-shot setting, and \textbf{108} training samples in Spanish and \textbf{64} training samples in Thai on the 3\% few-shot setting. Our models are trained on GTX 1080 Ti. The number of parameters for our models is around 5 million.

\subsection{Baselines}
We compare our model to the following baselines.

\paragraph{BiLSTM-CRF} This is the same cross-lingual SLU model structure as~\citet{schuster2019cross}.

\paragraph{BiLSTM-LVM} We replace the conditional random field (CRF) in BiLSTM-CRF with the LVM proposed in~\citet{liu2019zero}.





\paragraph{Multi. CoVe} 
Multilingual CoVe~\cite{yu-etal-2018-multilingual} is a bidirectional machine translation system that tends to encode phrases with similar meanings into similar vector spaces across languages. \citet{schuster2019cross} used it for the cross-lingual SLU task.

\paragraph{Multi. CoVe w/ auto-encoder} Based on Multilingual CoVe, \citet{schuster2019cross} added an auto-encoder objective so as to produce better-aligned representations for semantically similar sentences across languages.

\paragraph{Multilingual BERT (M-BERT)} It is a single language model pre-trained from monolingual corpora in 104 languages~\cite{devlin2019bert}, which is surprisingly good at cross-lingual model transfer.

\paragraph{Mixed Language Training (MLT)} \citet{liu2019attention} utilized keyword pairs to generate mixed language sentences for training cross-lingual task-oriented dialogue systems, which achieves promising zero-shot transfer ability.

\paragraph{CoSDA-ML} ~\citet{qin2020cosda} proposed a multilingual code-switching data augmentation framework to enhance the cross-lingual systems based on M-BERT~\cite{devlin2019bert}. It is a concurrent work of this paper.

\paragraph{XL-SLU} It is a previous state-of-the-art model in the zero-shot cross-lingual SLU task, which combines Gaussian noise, cross-lingual embeddings refinement, and the LVM~\cite{liu2019zero}.

\paragraph{Translate Train} \citet{schuster2019cross} trained a supervised machine translation system to translate English data into the target language, and then trained the model on the translated dataset.

\paragraph{All-shot Settings} We train the BiLSTM-CRF model~\cite{lample2016neural} on all the target language training samples, and on both the source and target language training set.

\section{Results \& Discussion}

\subsection{Few-shot Setting}
\paragraph{Quantitative Analysis}
The few-shot results are illustrated in Table~\ref{table:few-zero-shot}, from which we can clearly see consistent improvements made by label regularization and adversarial training. For example, on the 1\% few-shot setting, our model improves on BiLSTM-LVM in terms of accuracy/f1-score by 1.85\%/1.16\% in Spanish, and by 4.16\%/6.93\% in Thai. 
Our model also surpasses a strong baseline, M-BERT, while our model based on BiLSTM has many fewer parameters compared to M-BERT. For example, on the 1\% few-shot setting, our model improves on M-BERT in terms of accuracy/f1-score by 3.80\%/3.83\% in Thai.
Instead of generating a feature point like CRF, the LVM creates a more robust cross-lingual adaptation by generating a distribution for the intent or each token in the utterance. However, distributions generated by the LVM for the same slot type across languages might not be sufficiently close.
Incorporating adversarial training into the LVM alleviates this problem by regularizing the latent variables and making them more distinguishable. This improves the performance in both intent detection (a sentence-level task) and slot filling (a word-level task) by 0.92\%/3.16\% in Spanish and by 1.89\%/4.67\% in Thai on the 1\% few-shot setting. This proves that both sentence-level and word-level representations are better aligned across languages.

\begin{table}[!t]
\centering
\resizebox{0.42\textwidth}{!}{
\begin{tabular}{lcc}
\hline
\multicolumn{1}{c|}{\multirow{2}{*}{Model}}      & \multicolumn{2}{c}{\textbf{Thai}}                     \\ \cline{2-3} 
\multicolumn{1}{c|}{}                            & \textbf{Intent}           & \textbf{Slot}             \\ \hline
\multicolumn{3}{l}{\textit{few-shot on 5\% target language training set}}                                         \\ \hline
\multicolumn{1}{l|}{BiLSTM-CRF}                  & 90.05                     & 72.11                     \\
\multicolumn{1}{l|}{+ LR}                        & 91.11                     & 73.71                     \\ \hline
\multicolumn{1}{l|}{BiLSTM-LVM}                  & 91.02                     & 73.11                     \\
\multicolumn{1}{l|}{+ LR}                        & 91.45                     & 75.18                     \\
\multicolumn{1}{l|}{+ ALVM}                 & 91.08                     & 74.67                     \\
\multicolumn{1}{l|}{+ LR \& ALVM}           & 91.58                     & 75.87                     \\
\multicolumn{1}{l|}{+ LR \& ALVM \& delex.} & \textbf{92.51}   & \textbf{77.03}                     \\ \hline
\multicolumn{1}{l|}{XL-SLU}  & 91.05  & 73.43   \\ 
\multicolumn{1}{l|}{M-BERT}  & 92.02  & 75.52   \\\hline
\end{tabular}
}
\caption{Results of few-shot learning on 5\% Thai training data, which are averaged over three runs. We make the training samples in Thai the same as the 3\% Spanish training samples (\textbf{108}).}
\label{table:th5per}
\end{table}

In addition, LR aims to further align the sentence-level representations of target language utterances into a semantically similar space of source language utterances. As a result, there are 0.93\%/2.82\% improvements in intent detection for Spanish/Thai on the 1\% few-shot setting after we add LR to BiLSTM-LVM. Interestingly, the performance gains are not only on the intent detection but also on the slot filling, with an improvement of 1.77\%/3.94\% in Spanish/Thai.
This is attributed to the fact that utterance representations are produced based on word-level representations from BiLSTM. Therefore, the alignment of word-level representations will be implicitly improved in this process. Furthermore, incorporating LR and ALVM further tackles the inherent difficulties for the cross-lingual adaptation and achieves the state-of-the-art few-shot performance. Notably, by only leveraging 3\% of target language training samples, the results of our best model are on par with the supervised training on all the target language training data.

\paragraph{Adaptation ability to unrelated languages}
From Table~\ref{table:few-zero-shot}, we observe impressive improvements in Thai, an unrelated language to English, by utilizing our proposed approaches, especially when the number of target language training samples is small. For example, compared to the BiLSTM-LVM, our best model significantly improves the accuracy and f1-score by $\sim$4\%/$\sim$7\% in intent detection and slot filling in Thai in the few-shot setting on 1\% data. Additionally, in the same setting, our model surpasses the strong baseline, M-BERT, in terms of accuracy and f1-score by $\sim$4\%.
This illustrates that our approaches provide strong adaptation robustness and are able to tackle the inherent adaptation difficulties to unrelated languages.

\begin{table}[!t]
\centering
\resizebox{0.475\textwidth}{!}{
\begin{tabular}{lcccc}
\hline
\multicolumn{1}{c|}{\multirow{2}{*}{Model}}      & \multicolumn{2}{c|}{\textbf{Spanish}} & \multicolumn{2}{c}{\textbf{Thai}} \\ \cline{2-5} 
\multicolumn{1}{c|}{}   & \textbf{Intent}  & \multicolumn{1}{c|}{\textbf{Slot}}  & \textbf{Intent} & \textbf{Slot}   \\ \hline
\multicolumn{5}{l}{\textit{few-shot on 1\% target language training set}} \\ \hline
\multicolumn{1}{l|}{Our Model}                  & 93.82       & \multicolumn{1}{c|}{78.46}          & 87.43  & 62.44     \\
\multicolumn{1}{l|}{w/o Pre-training}                        & 92.75     &  \multicolumn{1}{c|}{77.11}          & 86.29                     & 60.20  \\ \hline \hline
\multicolumn{5}{l}{\textit{few-shot on 3\% target language training set}} \\ \hline
\multicolumn{1}{l|}{Our Model}                  & 95.20       & \multicolumn{1}{c|}{84.19}          & 90.97  & 70.88     \\
\multicolumn{1}{l|}{w/o Pre-training}                        & 94.51   & \multicolumn{1}{c|}{82.83}          & 89.72                     & 69.66  \\ \hline \hline
\multicolumn{5}{l}{\textit{zero-shot setting}} \\ \hline
\multicolumn{1}{l|}{Our Model}                  & 92.31       & \multicolumn{1}{c|}{72.49}          & 75.77  & 33.28     \\
\multicolumn{1}{l|}{w/o Pre-training}                        & 91.02       & \multicolumn{1}{c|}{71.72}          & 75.18                     & 32.69  \\ \hline
\end{tabular}
}
\caption{Results of the ablation study for the label sequence encoder pre-training (averaged over three runs). Our model refers to the one that combines LR, ALVM and delex. with BiLSTM-LVM.}
\label{table:ablation}
\end{table}

\begin{figure*}[!ht]
\centering
\begin{subfigure}{.49\textwidth}
    \centering
    \includegraphics[scale=0.31]{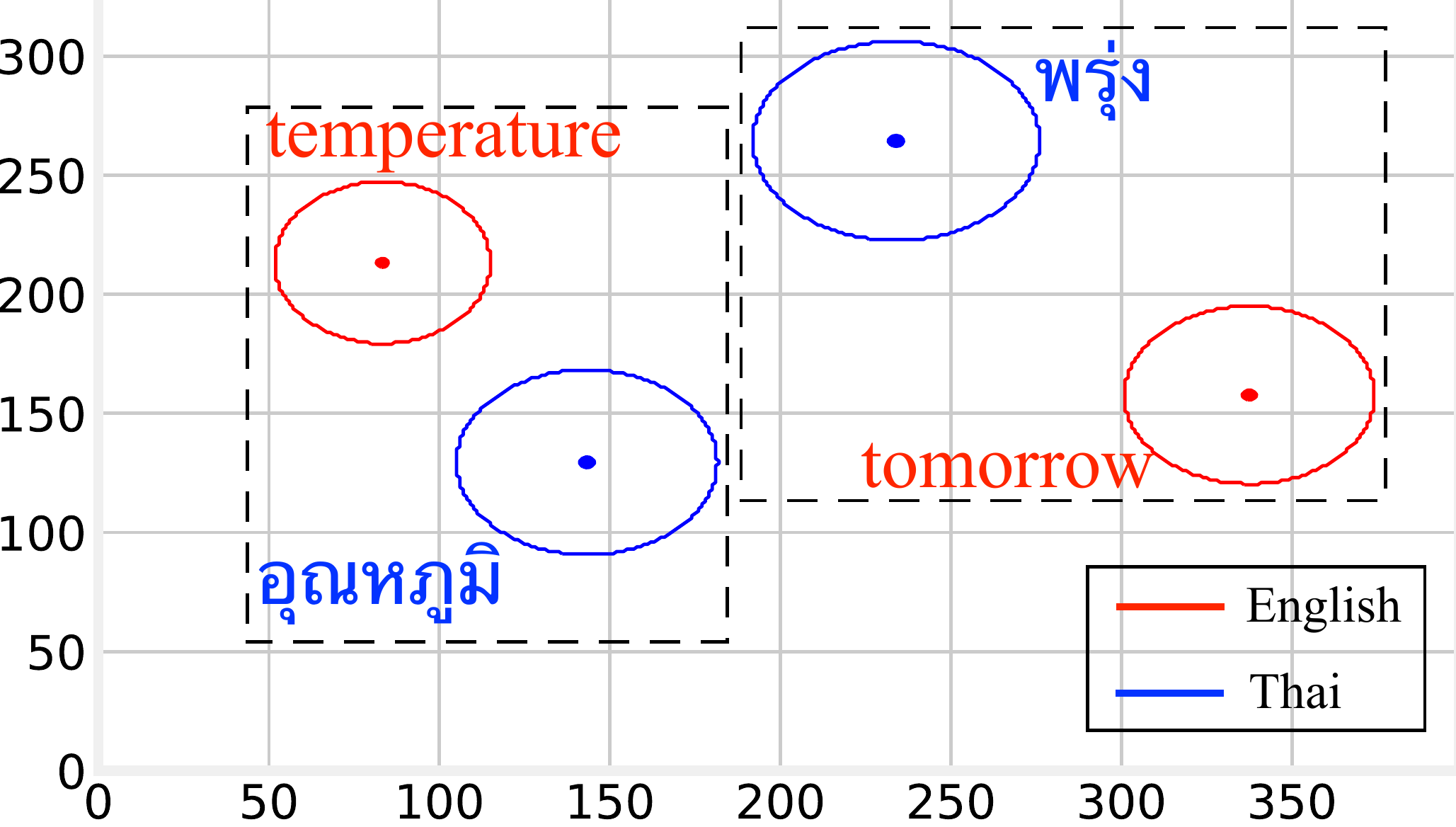}
    \caption{LVM}
    \label{fig:intent-es}
\end{subfigure}
\begin{subfigure}{.49\textwidth}
    \centering
    \includegraphics[scale=0.31]{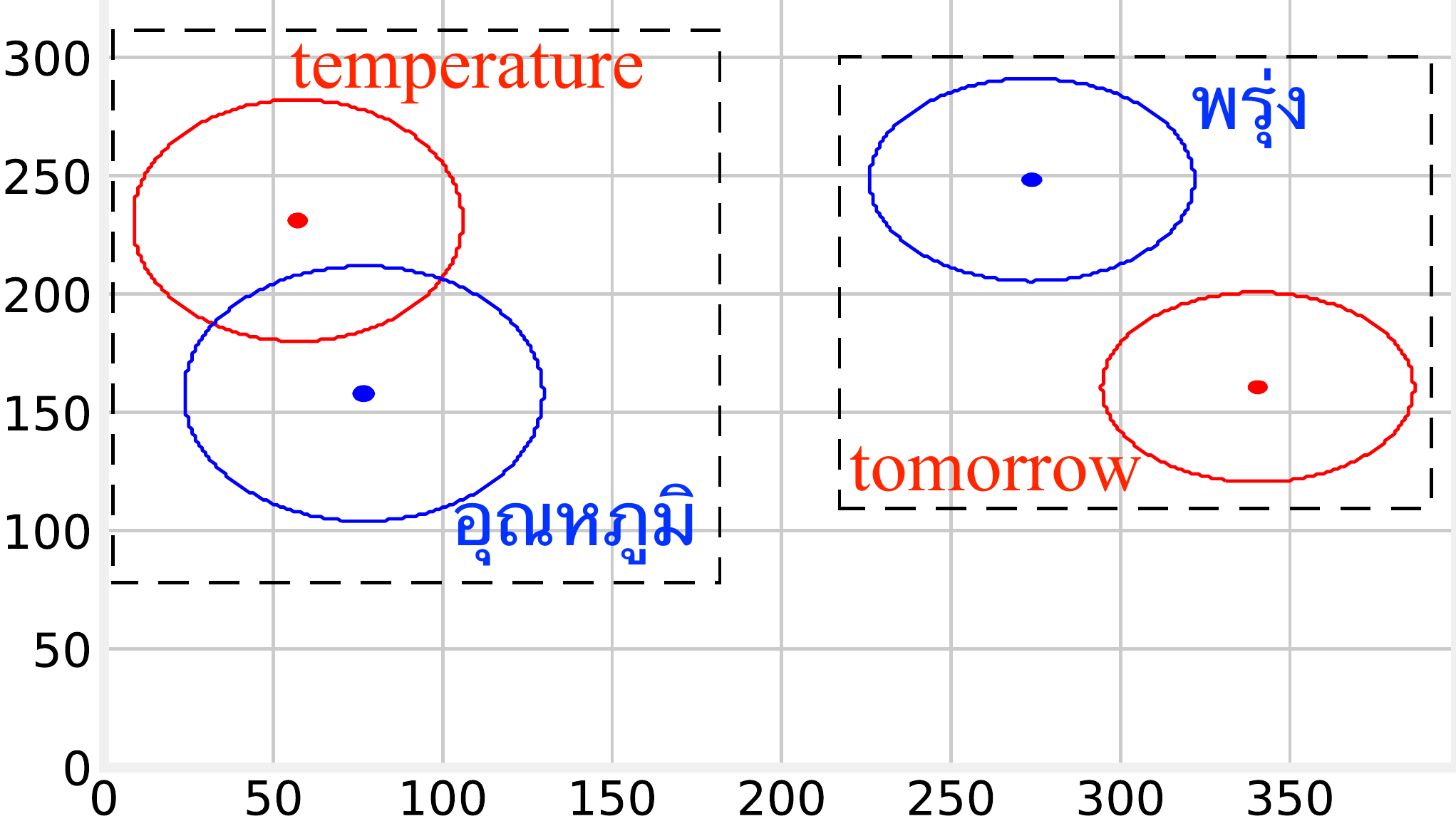}
    \caption{LVM + LR}
    \label{fig:slot-es}
\end{subfigure}
\begin{subfigure}{.49\textwidth}
    \centering
    \includegraphics[scale=0.31]{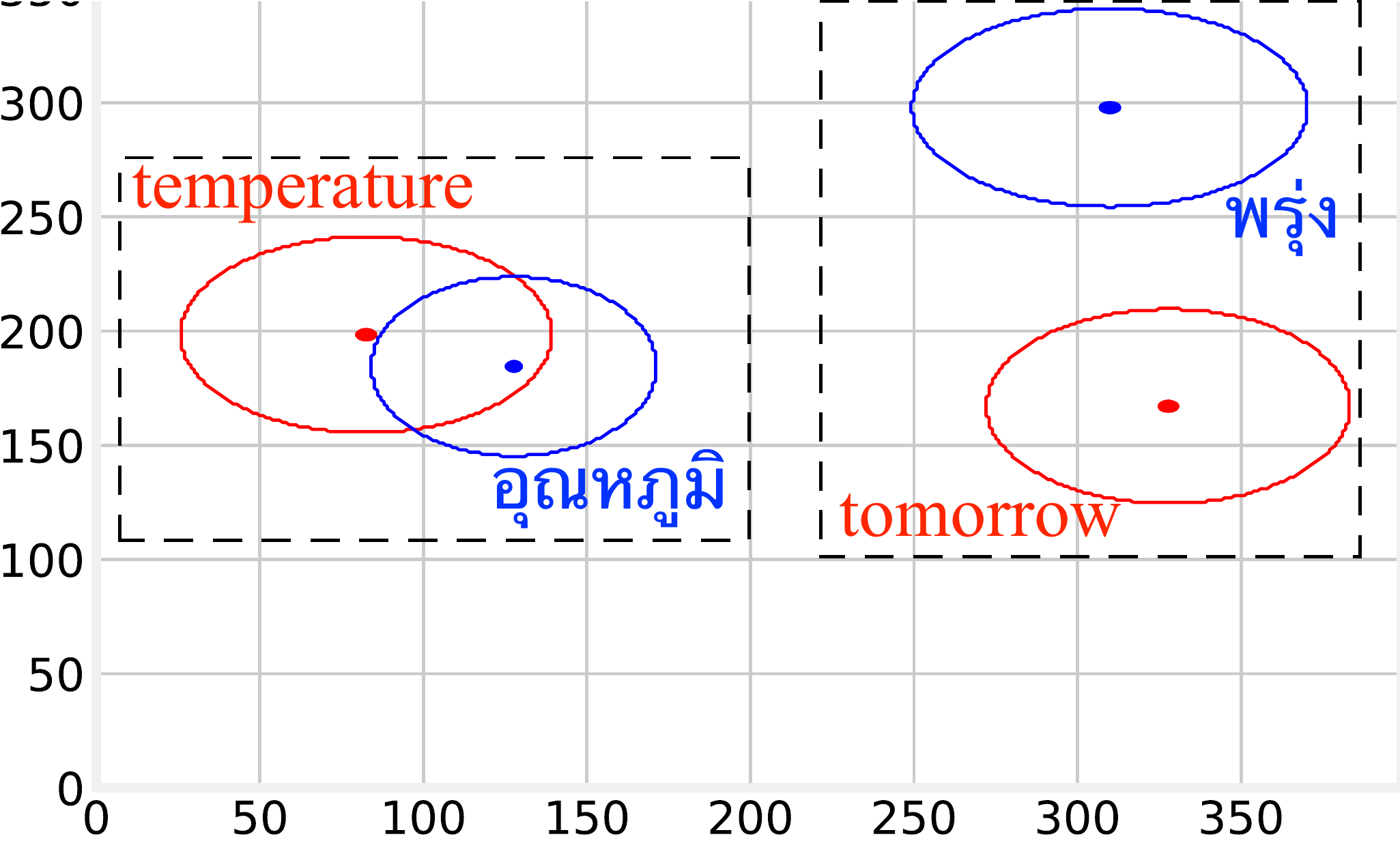}
    \caption{ALVM}
    \label{fig:intent-es-unseen}
\end{subfigure}
\begin{subfigure}{.49\textwidth}
    \centering
    \includegraphics[scale=0.31]{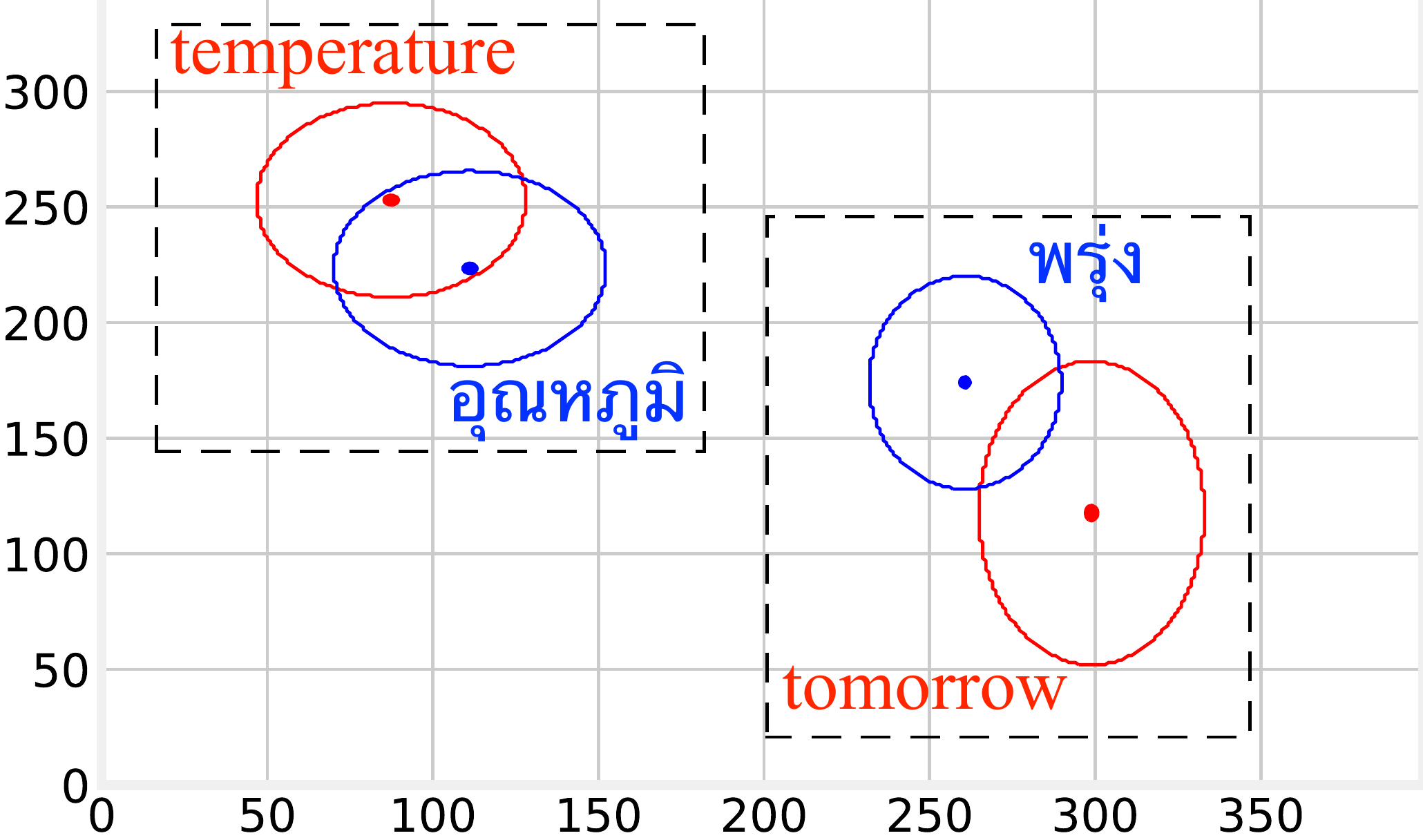}
    \caption{ALVM + LR}
    \label{fig:slot-es-unseen}
\end{subfigure}
\caption{Visualization for latent variables of parallel word pairs in English and Thai over different models trained on 1\% target language training set. We choose the word pairs ``temperature-
\foreignlanguage{thaicjk}{อุณหภูมิ}''
and ``tomorrow-
\foreignlanguage{thaicjk}{พรุ่ง}''
from the parallel sentences ``what will be the temperature tomorrow'' and ``\foreignlanguage{thaicjk}{อุณหภูมิ จะ อยู่ ท เท่า ไหร่ พรุ่ง}''
in English and Thai, respectively. To draw the contour plot, we sample 3000 points from the distribution of latent variables for the selected words, use PCA to project those points into 2D and calculate the mean and variance for each word.}
\label{fig:visualization}
\end{figure*}

\paragraph{Comparison between Spanish and Thai}
To make a fair comparison for the few-shot performance in Spanish and Thai, we increase the training size of Thai to the same as 3\% Spanish training samples, as depicted in Table~\ref{table:th5per}. We can see that there is still a performance gap between the Spanish and Thai (3.11\% in the intent detection task and 8.15\% in the slot filling task). This is because Spanish is grammatically and syntactically closer to English than Thai, leading to a better quality of cross-lingual alignment.

\paragraph{Visualization of Latent Variables}
The effectiveness of the LR and ALVM can be clearly seen from Figure~\ref{fig:visualization}. The former approach decreases the distance of latent variables for words with similar semantic meanings in different languages.
For the latter approach, to make the distributions for different slot types distinguishable, our model regularizes the latent variables of different slot types far from each other, and eventually it also improves the alignment of words with the same slot type. Incorporating both approaches further improves the word-level alignment across languages.
It further proves the robustness of our proposed approaches when adapting from the source language (English) to the unrelated language (Thai).

\subsection{Zero-shot Setting}
From Table~\ref{table:few-zero-shot}, we observe the remarkable improvements made by LR and ALVM on the state-of-the-art model XL-SLU in the zero-shot setting, and the slot filling performance of our best model in Spanish is on par with the strong baseline Translate Train, which leverages large amounts of bilingual resources. LR improves the adaptation robustness by making the word-level and sentence-level representations of similar utterances distinguishable. In addition, integrating adversarial training with the LVM further increases the robustness by disentangling the latent variables for different slot types. However, the performance boost for slot filling in Thai is limited. We conjecture that the inherent discrepancies in cross-lingual word embeddings and language structures for topologically different languages pairs make the word-level representations between them difficult to align in the zero-shot scenario. We notice that Multilingual CoVe with auto-encoder achieves slightly better performance than our model on the slot filling task in Thai. This is because this baseline leverages large amounts of monolingual and bilingual resources, which largely benefits the cross-lingual alignment between English and Thai. CoSDA-ML, a concurrent work of our model, utilizes additional augmented multilingual code-switching data, which significantly improves the zero-shot cross-lingual performance.

\subsection{Effectiveness of Label Sequence Encoder Pre-training}
Label sequence encoder pre-training helps the label encoder to generate more expressive representations for label sequences, which ensures the effectiveness of the label regularization approach.
From Table~\ref{table:ablation}, we can clearly observe the consistent performance gains made by pre-training in both few-shot and zero-shot scenarios. 

\section{Conclusion}
Current cross-lingual SLU models still suffer from imperfect cross-lingual alignments between the source and target languages.
In this paper, we propose label regularization (LR) and the adversarial latent variable model (ALVM) to regularize and further align the word-level and sentence-level representations across languages without utilizing any additional bilingual resources.
Experiments on the cross-lingual SLU task illustrate that our model achieves a remarkable performance boost compared to the strong baselines in both zero-shot and few-shot scenarios, and our model has a robust adaptation ability to unrelated target languages in the few-shot scenario. In addition, visualization for latent variables further proves that our approaches are effective at improving the alignment of cross-lingual representations.

\section*{Acknowledgments}
This work is partially funded by ITF/319/16FP and MRP/055/18 of the Innovation Technology Commission, the Hong Kong SAR Government.

\bibliographystyle{acl_natbib}
\bibliography{emnlp2020}

\begin{thebibliography}{33}
\expandafter\ifx\csname natexlab\endcsname\relax\def\natexlab#1{#1}\fi

\bibitem[{Artetxe et~al.(2017)Artetxe, Labaka, and
  Agirre}]{artetxe2017learning}
Mikel Artetxe, Gorka Labaka, and Eneko Agirre. 2017.
\newblock Learning bilingual word embeddings with (almost) no bilingual data.
\newblock In \emph{Proceedings of the 55th Annual Meeting of the Association
  for Computational Linguistics (Volume 1: Long Papers)}, pages 451--462.

\bibitem[{Bapna et~al.(2017)Bapna, T{\"u}r, Hakkani-T{\"u}r, and
  Heck}]{bapna2017towards}
Ankur Bapna, Gokhan T{\"u}r, Dilek Hakkani-T{\"u}r, and Larry Heck. 2017.
\newblock Towards zero-shot frame semantic parsing for domain scaling.
\newblock \emph{Proc. Interspeech 2017}, pages 2476--2480.

\bibitem[{Chen et~al.(2019)Chen, Zhuo, and Wang}]{chen2019bert}
Qian Chen, Zhu Zhuo, and Wen Wang. 2019.
\newblock Bert for joint intent classification and slot filling.
\newblock \emph{arXiv preprint arXiv:1902.10909}.

\bibitem[{Chen et~al.(2018)Chen, Chen, Su, Wang, Yu, Yan, and
  Wang}]{chen2018xl}
Wenhu Chen, Jianshu Chen, Yu~Su, Xin Wang, Dong Yu, Xifeng Yan, and
  William~Yang Wang. 2018.
\newblock Xl-nbt: A cross-lingual neural belief tracking framework.
\newblock In \emph{Proceedings of the 2018 Conference on Empirical Methods in
  Natural Language Processing}, pages 414--424.

\bibitem[{Conneau and Lample(2019)}]{lample2019cross}
Alexis Conneau and Guillaume Lample. 2019.
\newblock \href
  {http://papers.nips.cc/paper/8928-cross-lingual-language-model-pretraining.pdf}
  {Cross-lingual language model pretraining}.
\newblock In \emph{Advances in Neural Information Processing Systems 32}, pages
  7057--7067. Curran Associates, Inc.

\bibitem[{Conneau et~al.(2018)Conneau, Lample, Ranzato, Denoyer, and
  J\'egou}]{conneau2017word}
Alexis Conneau, Guillaume Lample, Marc'Aurelio Ranzato, Ludovic Denoyer, and
  Herv\'e J\'egou. 2018.
\newblock Word translation without parallel data.
\newblock In \emph{International Conference on Learning Representations
  (ICLR)}.

\bibitem[{Devlin et~al.(2019)Devlin, Chang, Lee, and
  Toutanova}]{devlin2019bert}
Jacob Devlin, Ming-Wei Chang, Kenton Lee, and Kristina Toutanova. 2019.
\newblock Bert: Pre-training of deep bidirectional transformers for language
  understanding.
\newblock In \emph{Proceedings of the 2019 Conference of the North American
  Chapter of the Association for Computational Linguistics: Human Language
  Technologies, Volume 1 (Long and Short Papers)}, pages 4171--4186.

\bibitem[{Felbo et~al.(2017)Felbo, Mislove, S{\o}gaard, Rahwan, and
  Lehmann}]{felbo2017using}
Bjarke Felbo, Alan Mislove, Anders S{\o}gaard, Iyad Rahwan, and Sune Lehmann.
  2017.
\newblock Using millions of emoji occurrences to learn any-domain
  representations for detecting sentiment, emotion and sarcasm.
\newblock In \emph{Proceedings of the 2017 Conference on Empirical Methods in
  Natural Language Processing}, pages 1615--1625.

\bibitem[{Goo et~al.(2018)Goo, Gao, Hsu, Huo, Chen, Hsu, and
  Chen}]{goo2018slot}
Chih-Wen Goo, Guang Gao, Yun-Kai Hsu, Chih-Li Huo, Tsung-Chieh Chen, Keng-Wei
  Hsu, and Yun-Nung Chen. 2018.
\newblock Slot-gated modeling for joint slot filling and intent prediction.
\newblock In \emph{Proceedings of the 2018 Conference of the North American
  Chapter of the Association for Computational Linguistics: Human Language
  Technologies, Volume 2 (Short Papers)}, pages 753--757.

\bibitem[{Goodfellow et~al.(2014)Goodfellow, Pouget-Abadie, Mirza, Xu,
  Warde-Farley, Ozair, Courville, and Bengio}]{goodfellow2014generative}
Ian Goodfellow, Jean Pouget-Abadie, Mehdi Mirza, Bing Xu, David Warde-Farley,
  Sherjil Ozair, Aaron Courville, and Yoshua Bengio. 2014.
\newblock Generative adversarial nets.
\newblock In \emph{Advances in neural information processing systems}, pages
  2672--2680.

\bibitem[{Haihong et~al.(2019)Haihong, Niu, Chen, and Song}]{haihong2019novel}
E~Haihong, Peiqing Niu, Zhongfu Chen, and Meina Song. 2019.
\newblock A novel bi-directional interrelated model for joint intent detection
  and slot filling.
\newblock In \emph{Proceedings of the 57th Annual Meeting of the Association
  for Computational Linguistics}, pages 5467--5471.

\bibitem[{Hochreiter and Schmidhuber(1997)}]{hochreiter1997long}
Sepp Hochreiter and J{\"u}rgen Schmidhuber. 1997.
\newblock Long short-term memory.
\newblock \emph{Neural Computation}, 9(8):1735--1780.

\bibitem[{Huang et~al.(2019)Huang, Liang, Duan, Gong, Shou, Jiang, and
  Zhou}]{huang2019unicoder}
Haoyang Huang, Yaobo Liang, Nan Duan, Ming Gong, Linjun Shou, Daxin Jiang, and
  Ming Zhou. 2019.
\newblock Unicoder: A universal language encoder by pre-training with multiple
  cross-lingual tasks.
\newblock In \emph{Proceedings of the 2019 Conference on Empirical Methods in
  Natural Language Processing and the 9th International Joint Conference on
  Natural Language Processing (EMNLP-IJCNLP)}, pages 2485--2494.

\bibitem[{Kim et~al.(2017)Kim, Kim, Sarikaya, and
  Fosler-Lussier}]{kim2017cross}
Joo-Kyung Kim, Young-Bum Kim, Ruhi Sarikaya, and Eric Fosler-Lussier. 2017.
\newblock Cross-lingual transfer learning for pos tagging without cross-lingual
  resources.
\newblock In \emph{Proceedings of the 2017 Conference on Empirical Methods in
  Natural Language Processing}, pages 2832--2838.

\bibitem[{Lample et~al.(2016)Lample, Ballesteros, Subramanian, Kawakami, and
  Dyer}]{lample2016neural}
Guillaume Lample, Miguel Ballesteros, Sandeep Subramanian, Kazuya Kawakami, and
  Chris Dyer. 2016.
\newblock Neural architectures for named entity recognition.
\newblock In \emph{Proceedings of the 2016 Conference of the North American
  Chapter of the Association for Computational Linguistics: Human Language
  Technologies}, pages 260--270.

\bibitem[{Lin et~al.(2020)Lin, Liu, Winata, Cahyawijaya, Madotto, Bang, Ishii,
  and Fung}]{lin2020xpersona}
Zhaojiang Lin, Zihan Liu, Genta~Indra Winata, Samuel Cahyawijaya, Andrea
  Madotto, Yejin Bang, Etsuko Ishii, and Pascale Fung. 2020.
\newblock Xpersona: Evaluating multilingual personalized chatbot.
\newblock \emph{arXiv preprint arXiv:2003.07568}.

\bibitem[{Liu et~al.(2019{\natexlab{a}})Liu, Shin, Xu, Winata, Xu, Madotto, and
  Fung}]{liu2019zero}
Zihan Liu, Jamin Shin, Yan Xu, Genta~Indra Winata, Peng Xu, Andrea Madotto, and
  Pascale Fung. 2019{\natexlab{a}}.
\newblock Zero-shot cross-lingual dialogue systems with transferable latent
  variables.
\newblock In \emph{Proceedings of the 2019 Conference on Empirical Methods in
  Natural Language Processing and the 9th International Joint Conference on
  Natural Language Processing (EMNLP-IJCNLP)}, pages 1297--1303.

\bibitem[{Liu et~al.(2019{\natexlab{b}})Liu, Winata, Lin, Xu, and
  Fung}]{liu2019attention}
Zihan Liu, Genta~Indra Winata, Zhaojiang Lin, Peng Xu, and Pascale Fung.
  2019{\natexlab{b}}.
\newblock Attention-informed mixed-language training for zero-shot
  cross-lingual task-oriented dialogue systems.
\newblock \emph{arXiv preprint arXiv:1911.09273}.

\bibitem[{Liu et~al.(2020{\natexlab{a}})Liu, Winata, Madotto, and
  Fung}]{liu2020exploring}
Zihan Liu, Genta~Indra Winata, Andrea Madotto, and Pascale Fung.
  2020{\natexlab{a}}.
\newblock Exploring fine-tuning techniques for pre-trained cross-lingual models
  via continual learning.
\newblock \emph{arXiv preprint arXiv:2004.14218}.

\bibitem[{Liu et~al.(2020{\natexlab{b}})Liu, Winata, Xu, and
  Fung}]{liu-etal-2020-coach}
Zihan Liu, Genta~Indra Winata, Peng Xu, and Pascale Fung. 2020{\natexlab{b}}.
\newblock \href {https://doi.org/10.18653/v1/2020.acl-main.3} {{C}oach: A
  coarse-to-fine approach for cross-domain slot filling}.
\newblock In \emph{Proceedings of the 58th Annual Meeting of the Association
  for Computational Linguistics}, pages 19--25, Online. Association for
  Computational Linguistics.

\bibitem[{Mayhew et~al.(2017)Mayhew, Tsai, and Roth}]{mayhew2017cheap}
Stephen Mayhew, Chen-Tse Tsai, and Dan Roth. 2017.
\newblock Cheap translation for cross-lingual named entity recognition.
\newblock In \emph{Proceedings of the 2017 conference on empirical methods in
  natural language processing}, pages 2536--2545.

\bibitem[{Mrk{\v{s}}i{\'c} et~al.(2017)Mrk{\v{s}}i{\'c}, Vuli{\'c},
  S{\'e}aghdha, Leviant, Reichart, Ga{\v{s}}i{\'c}, Korhonen, and
  Young}]{mrkvsic2017semantic}
Nikola Mrk{\v{s}}i{\'c}, Ivan Vuli{\'c}, Diarmuid~{\'O} S{\'e}aghdha, Ira
  Leviant, Roi Reichart, Milica Ga{\v{s}}i{\'c}, Anna Korhonen, and Steve
  Young. 2017.
\newblock Semantic specialization of distributional word vector spaces using
  monolingual and cross-lingual constraints.
\newblock \emph{Transactions of the Association for Computational Linguistics},
  5(1):309--324.

\bibitem[{Pires et~al.(2019)Pires, Schlinger, and
  Garrette}]{pires-etal-2019-multilingual}
Telmo Pires, Eva Schlinger, and Dan Garrette. 2019.
\newblock \href {https://doi.org/10.18653/v1/P19-1493} {How multilingual is
  multilingual {BERT}?}
\newblock In \emph{Proceedings of the 57th Annual Meeting of the Association
  for Computational Linguistics}, pages 4996--5001, Florence, Italy.
  Association for Computational Linguistics.

\bibitem[{Qin et~al.(2020)Qin, Ni, Zhang, and Che}]{qin2020cosda}
Libo Qin, Minheng Ni, Yue Zhang, and Wanxiang Che. 2020.
\newblock Cosda-ml: Multi-lingual code-switching data augmentation for
  zero-shot cross-lingual nlp.
\newblock \emph{arXiv preprint arXiv:2006.06402}.

\bibitem[{Schuster et~al.(2019)Schuster, Gupta, Shah, and
  Lewis}]{schuster2019cross}
Sebastian Schuster, Sonal Gupta, Rushin Shah, and Mike Lewis. 2019.
\newblock Cross-lingual transfer learning for multilingual task oriented
  dialog.
\newblock In \emph{Proceedings of the 2019 Conference of the North American
  Chapter of the Association for Computational Linguistics: Human Language
  Technologies, Volume 1 (Long and Short Papers)}, pages 3795--3805.

\bibitem[{Sil et~al.(2018)Sil, Kundu, Florian, and Hamza}]{sil2018neural}
Avirup Sil, Gourab Kundu, Radu Florian, and Wael Hamza. 2018.
\newblock Neural cross-lingual entity linking.
\newblock In \emph{Thirty-Second AAAI Conference on Artificial Intelligence}.

\bibitem[{Upadhyay et~al.(2018{\natexlab{a}})Upadhyay, Faruqui, T{\"u}r, Dilek,
  and Heck}]{upadhyay2018almost}
Shyam Upadhyay, Manaal Faruqui, Gokhan T{\"u}r, Hakkani-T{\"u}r Dilek, and
  Larry Heck. 2018{\natexlab{a}}.
\newblock (almost) zero-shot cross-lingual spoken language understanding.
\newblock In \emph{2018 IEEE International Conference on Acoustics, Speech and
  Signal Processing (ICASSP)}, pages 6034--6038. IEEE.

\bibitem[{Upadhyay et~al.(2018{\natexlab{b}})Upadhyay, Gupta, and
  Roth}]{upadhyay2018joint}
Shyam Upadhyay, Nitish Gupta, and Dan Roth. 2018{\natexlab{b}}.
\newblock Joint multilingual supervision for cross-lingual entity linking.
\newblock In \emph{Proceedings of the 2018 Conference on Empirical Methods in
  Natural Language Processing}, pages 2486--2495.

\bibitem[{Wu et~al.(2019)Wu, Madotto, Hosseini-Asl, Xiong, Socher, and
  Fung}]{wu2019transferable}
Chien-Sheng Wu, Andrea Madotto, Ehsan Hosseini-Asl, Caiming Xiong, Richard
  Socher, and Pascale Fung. 2019.
\newblock Transferable multi-domain state generator for task-oriented dialogue
  systems.
\newblock In \emph{Proceedings of the 57th Annual Meeting of the Association
  for Computational Linguistics}, pages 808--819.

\bibitem[{Xie et~al.(2018)Xie, Yang, Neubig, Smith, and
  Carbonell}]{xie2018neural}
Jiateng Xie, Zhilin Yang, Graham Neubig, Noah~A Smith, and Jaime Carbonell.
  2018.
\newblock Neural cross-lingual named entity recognition with minimal resources.
\newblock In \emph{Proceedings of the 2018 Conference on Empirical Methods in
  Natural Language Processing}, pages 369--379.

\bibitem[{Yu et~al.(2018)Yu, Li, and Oguz}]{yu-etal-2018-multilingual}
Katherine Yu, Haoran Li, and Barlas Oguz. 2018.
\newblock Multilingual seq2seq training with similarity loss for cross-lingual
  document classification.
\newblock In \emph{Proceedings of The Third Workshop on Representation Learning
  for {NLP}}, pages 175--179, Melbourne, Australia. Association for
  Computational Linguistics.

\bibitem[{Zhang et~al.(2013)Zhang, Liu, and Zhao}]{zhang2013cross}
Tao Zhang, Kang Liu, and Jun Zhao. 2013.
\newblock Cross lingual entity linking with bilingual topic model.
\newblock In \emph{Twenty-Third International Joint Conference on Artificial
  Intelligence}.

\bibitem[{Zhang et~al.(2016)Zhang, Gaddy, Barzilay, and
  Jaakkola}]{zhang2016ten}
Yuan Zhang, David Gaddy, Regina Barzilay, and Tommi Jaakkola. 2016.
\newblock Ten pairs to tag--multilingual pos tagging via coarse mapping between
  embeddings.
\newblock In \emph{Proceedings of the 2016 Conference of the North American
  Chapter of the Association for Computational Linguistics: Human Language
  Technologies}, pages 1307--1317.

\end{thebibliography}

\end{document}